%% file: AnonymousSubmission2027.tex
\documentclass[letterpaper]{article}

\usepackage[preprint]{aaai2027}
\nocopyright
\usepackage[hyphens]{url}
\usepackage{graphicx}
\usepackage{caption}
\usepackage{booktabs}
\usepackage{tabularx}
\usepackage{multirow}
\usepackage{amsmath}
\usepackage{amssymb}
\usepackage{amsthm}
\usepackage{pifont}
\usepackage{xcolor}
\usepackage[capitalise,noabbrev]{cleveref}
\usepackage{natbib}

\usepackage{algorithm}
\usepackage{algorithmic}
\usepackage{newfloat}
\usepackage{listings}
\usepackage[most]{tcolorbox}

\DeclareCaptionStyle{ruled}{
    labelfont=normalfont,
    labelsep=colon,
    strut=off
}

\definecolor{PromptHeader}{HTML}{A3A3A3}
\definecolor{PromptFrame}{HTML}{9C9C9C}
\newtcblisting{promptbox}[1]{
    enhanced jigsaw,
    breakable,
    lines before break=10,
    listing only,
    listing engine=listings,
    title={#1},
    title after break={#1\space (continued)},
    colback=white,
    colframe=PromptFrame,
    colbacktitle=PromptHeader,
    coltitle=white,
    fonttitle=\large\bfseries,
    boxrule=0.8pt,
    arc=2.5mm,
    outer arc=2.5mm,
    left=6mm,
    right=6mm,
    top=4mm,
    bottom=4mm,
    boxsep=0pt,
    toptitle=1.6mm,
    bottomtitle=1.6mm,
    lefttitle=5mm,
    before skip=10pt,
    after skip=12pt,
    listing options={
        basicstyle=\small\ttfamily,
        numbers=none,
        showstringspaces=false,
        columns=fullflexible,
        keepspaces=true,
        breaklines=true,
        breakatwhitespace=false,
        breakindent=1em,
        aboveskip=0pt,
        belowskip=0pt
    }
}

\newtcblisting{promptcontinuation}{
    enhanced jigsaw,
    listing only,
    listing engine=listings,
    colback=white,
    colframe=PromptFrame,
    boxrule=0.8pt,
    arc=2.5mm,
    outer arc=2.5mm,
    left=6mm,
    right=6mm,
    top=4mm,
    bottom=4mm,
    boxsep=0pt,
    before skip=10pt,
    after skip=12pt,
    listing options={
        basicstyle=\small\ttfamily,
        numbers=none,
        showstringspaces=false,
        columns=fullflexible,
        keepspaces=true,
        breaklines=true,
        breakatwhitespace=false,
        breakindent=1em,
        aboveskip=0pt,
        belowskip=0pt
    }
}

\floatstyle{ruled}
\newfloat{listing}{tb}{lst}{}
\floatname{listing}{Listing}

\theoremstyle{plain}

\newcolumntype{C}[1]{>{\centering\arraybackslash}p{#1}}

\makeatletter
\def\hlinew#1{%
    \noalign{\ifnum0=`}\fi
    \hrule \@height #1
    \futurelet\reserved@a\@xhline
}
\makeatother

\title{
Does Forgetting Transfer Across Modalities? A Real-World Benchmark for Cross-Modal Knowledge Unlearning Evaluation
}

\author{
Chunlin Liu\textsuperscript{1,*},
Junnian Chen\textsuperscript{1,*},
Haitong Jiang\textsuperscript{1},
Jianyu Zhao\textsuperscript{2},
Yingsen Pang\textsuperscript{1},
Jingchen Li\textsuperscript{1},
Jiabiao He\textsuperscript{1},
Youming Lu\textsuperscript{1},
Jinhe Bi\textsuperscript{3},
Yuntao Du\textsuperscript{4,$\dagger$}
}
\affiliations{
\textsuperscript{1}Shenzhen University\\
\textsuperscript{2}Hebei University of Technology\\
\textsuperscript{3}Ludwig Maximilian University of Munich\\
\textsuperscript{4}Shandong University\\
\textsuperscript{*}Equal contribution, \textsuperscript{$\dagger$}Corresponding author
}

\begin{document}

\maketitle

% ==================== 摘要 ====================
\begin{abstract}
Vision-Language Models (VLMs), like Large Language Models (LLMs), may memorize sensitive, copyrighted, or harmful knowledge from their pretraining corpora. Removing such knowledge is essential for building trustworthy AI systems. However, existing studies primarily focus on forgetting within individual modalities. Although recent work has begun to explore cross-modal consistency in unlearning, the cross-modal transfer of real-world knowledge unlearning remains insufficiently studied. To address this gap, we introduce \textbf{UNLINK-VL}, a real-world benchmark for cross-modal knowledge unlearning in VLMs. Under a post-hoc unlearning setting in which the original forget and retain corpora are unavailable, UNLINK-VL selects visually identifiable real-world entities as unlearning targets and associates them with corresponding images and one-hop and multi-hop facts derived from Wikidata. The benchmark comprises four complementary subsets that evaluate direct forgetting of target knowledge, the propagation of forgetting through relational knowledge, the preservation of related non-target knowledge, and robustness to semantically equivalent queries. We train models under text-only and multimodal unlearning settings and evaluate forgetting effectiveness and retained utility across textual, visual, and cross-modal scenarios. Extensive experiments reveal a pronounced asymmetry in cross-modal transfer: multimodal unlearning remains effective under textual evaluation, whereas text-only unlearning transfers poorly to visual and cross-modal scenarios. Meanwhile, the evaluated methods largely preserve the models’ general capabilities. These findings demonstrate that relying solely on intra-modal evaluation, particularly text-only evaluation, may substantially overestimate the effectiveness of knowledge unlearning in VLMs, underscoring the need for cross-modal unlearning and evaluation.
\end{abstract}

% 视觉语言模型（VLMs）可能会从多模态预训练语料中记忆敏感、
% 受版权保护或有害的知识。对这类知识进行遗忘消除比在纯文本
% 模型中更具挑战性，因为目标知识可通过文本查询、视觉输入或
% 跨模态推理被触发还原。
%
% 为此，我们提出 UNLINK-VL，一套适用于事后遗忘实用场景的、
% 面向视觉语言模型跨模态知识遗忘消除的真实数据集基准。
% 在该场景中，原始待遗忘语料与保留语料均无法获取。
%
% UNLINK-VL 选取可通过视觉识别的现实实体作为遗忘目标，
% 采集对应实体图像，并从维基数据中提取相关单跳与多跳事实。
% 针对每个目标实体，我们额外选取两个语义相近但关联知识
% 相互独立的实体，以此实现对邻近知识附带损伤的系统化评估。
%
% 依托这套结构化知识，UNLINK-VL 构建了四套互补评测子集，
% 分别用于评估：
% 1. 直接遗忘效果；
% 2. 遗忘效应在关联知识中的传递程度；
% 3. 非目标关联知识的留存能力；
% 4. 模型面对语义等价查询时的遗忘鲁棒性。
%
% 该基准还配套成对的正向、拒答与负向回复样本，
% 可支撑 GA、DPO、NPO、RT 等主流代表性遗忘消除方法开展实验。
%
% 我们在纯文本与跨模态场景下，对遗忘效果和模型保留性能进行评测。
% 大量实验揭示，跨模态迁移存在明显不对称性：多模态遗忘消除方法在文本评测环境下仍能保持有效，
% 但仅基于文本的遗忘消除方案很难适配视觉与跨模态场景。
%
% 同时，所有被测方法均可保留模型绝大多数通用基础能力。
% 上述结果表明，仅采用纯文本评测会大幅高估视觉语言模型
% 知识遗忘消除的真实效果，也凸显了开展跨模态评测的必要性。

% ==================== 正文 ====================
\input{Sec/1_intro}
\input{Sec/2_related_work}
\input{Sec/3_datasets}

\input{Sec/4_exp}
\input{Sec/5_conclusion}

% \newpage
\bibliography{aaai2027}

% ==================== 附录 ====================
% \clearpage
% \input{Sec/6_appx}

% ==================== 可复现性检查表 ====================
% \input{ReproducibilityChecklist}

\end{document}

%% file: Sec/1_intro.tex
\section{Introduction}
\label{sec:intro}

\begin{figure}[!t]
    \centering
    \includegraphics[width=\columnwidth]{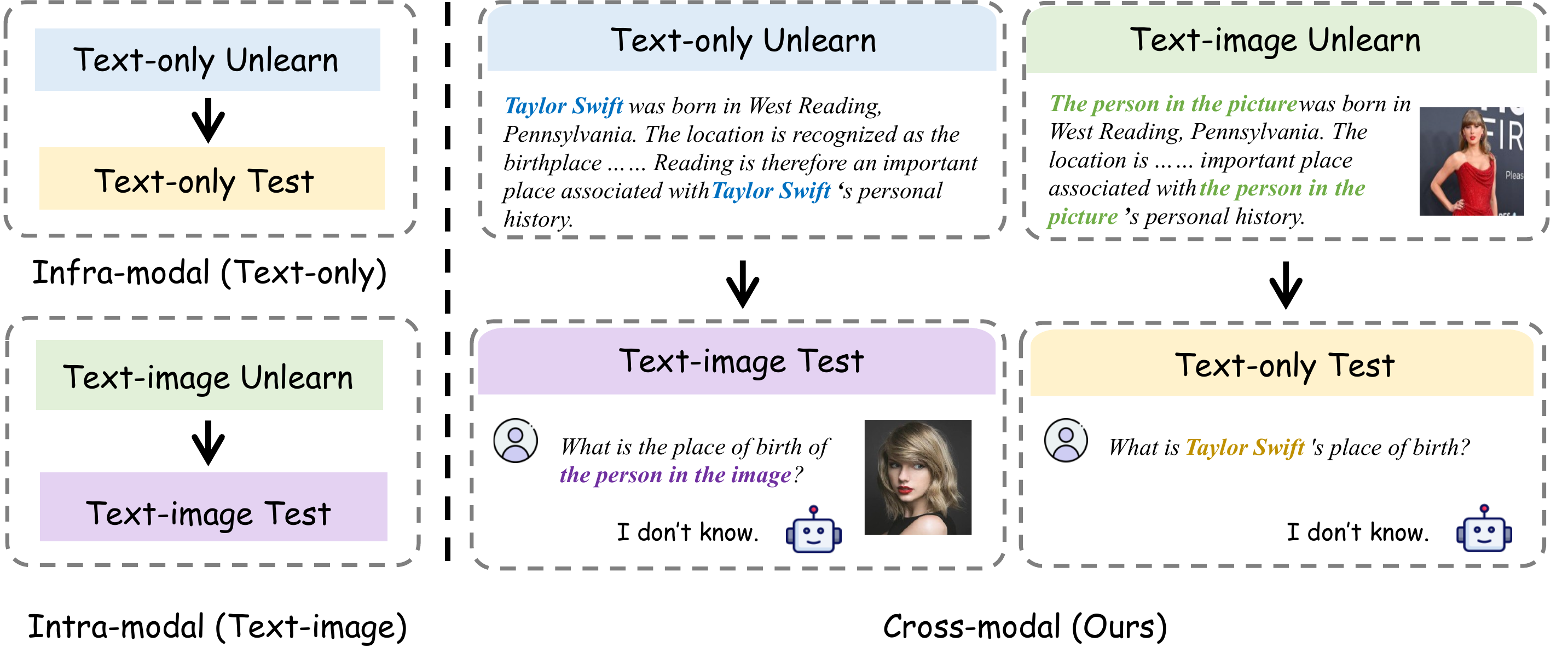}
    \caption{Comparison between intra-modal and cross-modal knowledge unlearning.}
    \label{fig:pipeline_overview}
    \vspace{-3mm}
\end{figure}

Vision-Language Models (VLMs), like Large Language Models (LLMs), learn from large-scale pretraining datasets and may memorize sensitive information, copyrighted content, or harmful knowledge~\cite{carlini2021extracting,ju2025album,liu2025mllmubench,zheng2025offside}. Once encoded in model parameters, such information or knowledge may later be elicited or reproduced during downstream deployment, posing privacy, copyright, and safety risks. Retraining a model from scratch after removing the target data from the training datasets is often prohibitively expensive~\cite{bourtoule2021machine,yao2024machine}. Machine unlearning seeks to remove the influence of designated training data or knowledge from a trained model while preserving non-target knowledge and general capabilities~\cite{ginart2019making,guo2020certified,bourtoule2021machine}.

Several benchmarks have been proposed to evaluate knowledge unlearning. Text-only benchmarks evaluate the forgetting of synthetic profiles, hazardous knowledge, and real-world entities using text-only inputs for both unlearning and evaluation~\cite{maini2024tofu,li2024wmdp,jin2024rwku}. Multimodal benchmarks extend evaluation to text-image inputs, facial identities, visual concepts, misinformation, and sensitive associations~\cite{liu2025mllmubench,dontsov2025clear,ma2025fiubench,li2024siu,zheng2025offside,selvassala2026salmubench}. Existing works mainly investigate intra-modal unlearning, where the model is trained to forget knowledge in one modality and evaluated in the same modality. Despite the progress enabled by these benchmarks, they largely overlook cross-modal transfer of the unlearning effect, which is a realistic and critical challenge in real-world multimodal applications.

For VLMs, robust knowledge unlearning should remain effective under both intra-modal and cross-modal evaluation. The same fact can be elicited through semantically equivalent text-only and text-image inputs. For example, a person may be specified by name in a text-only question or represented by an image in a text-image question, with both queries targeting the same biographical fact. A model may appear to have forgotten that fact when queried by name yet still recover it when shown the person's image. The unlearning effect should therefore exhibit cross-modal consistency, regardless of how unlearning is performed.

Recent benchmarks have begun to explore whether unlearning remains effective across different modality settings. UMU-Bench targets modality misalignment by evaluating unimodal, multimodal, and hybrid unlearning under both unimodal and multimodal settings~\cite{wang2025umubench}. PPU-Bench evaluates complete, selective, and personalized unlearning on pre-existing knowledge about real-world public figures using paired text-only QA and multimodal VQA samples~\cite{guang2026ppubench}. However, these benchmarks either focus primarily on synthetic profile knowledge or do not treat the unlearning modality as an independent experimental variable.

In this work, we study the cross-modal transfer problem: if a fact is unlearned through text-only inputs, can it still transfer to a semantically equivalent text-image question, and if it is unlearned through a text-image input, can it still transfer to an equivalent text-only question? To answer these questions, one could evaluate $T \rightarrow M$ and $M \rightarrow T$ as cross-modal transfer settings, where $T$ and $M$ denote text-only and text-image inputs, respectively. We also report $T \rightarrow T$ and $M \rightarrow M$ as intra-modal reference settings. This directionality matters in practice because a deletion request may identify target knowledge in one modality, whereas later users may query the same knowledge through another.

To this end, we introduce \textbf{UNLINK-VL}, a benchmark grounded in real-world knowledge for evaluating cross-modal knowledge unlearning in VLMs. UNLINK-VL selects a diverse set of visually identifiable real-world entities as unlearning targets, each associated with factual knowledge already accessible to the original model. For each entity, we first collect and filter relevant one-hop facts and two-hop paths from Wikidata~\cite{vrandecic2014wikidata}. We then verbalize this structured knowledge into text-only passages for unlearning and construct text-only QA pairs for evaluation. For both data types, we create aligned text-image counterparts by replacing the entity name with an image-based reference and pairing the resulting text with representative images.

For unlearning, UNLINK-VL constructs four forms of supervision tailored to representative unlearning methods, including GA, DPO, NPO, and RT~\cite{jang2023knowledge,rafailov2023dpo,zhang2024npo}. Faithful positive passages are verbalized directly from the collected Wikidata facts, while counterfactual negative passages are generated by modifying factual claims in the positive passages. GA and NPO suppress the faithful positive passages, while DPO uses paired counterfactual negative and faithful positive passages. For RT, each target query is paired with a generic refusal response, such as ``I don't know,'' to train the model to abstain from answering questions about the target knowledge.

For evaluation, UNLINK-VL organizes the structured knowledge into four complementary subsets. The Forget Set evaluates direct forgetting of target facts; the Multi-Hop Set examines whether forgetting extends through two-hop relational paths; the Retain Set evaluates the preservation of non-target facts; and the Rewrite Set assesses whether forgetting remains effective under semantically equivalent queries. Each evaluated fact is represented in aligned text-only and text-image forms. To prevent leakage, the unlearning and evaluation splits use disjoint images.

We conduct extensive experiments with four representative unlearning methods on multiple VLMs. The results reveal a pronounced asymmetry in cross-modal transfer: multimodal unlearning transfers more effectively to text-only evaluation than text-only unlearning transfers to text-image evaluation. Meanwhile, the evaluated methods largely preserve general capabilities. These findings show that text-only evaluation may substantially overestimate unlearning effectiveness in VLMs and highlight the need to consider both text-only and text-image inputs during unlearning and evaluation.

In summary, our main contributions are:
\begin{itemize}
    \item We introduce \textbf{UNLINK-VL}, a real-world benchmark for evaluating the cross-modal transfer of unlearning effects in VLMs.
    \item We develop a systematic pipeline that constructs method-specific unlearning data and aligned text-only and text-image evaluation samples from pre-existing model knowledge, enabling comprehensive evaluation of forgetting, robustness, and knowledge preservation.
    \item Through controlled experiments that independently vary the unlearning and evaluation modalities, we reveal a pronounced asymmetry in cross-modal transfer and show that the evaluated methods largely preserve general capabilities.
\end{itemize}

%% file: Sec/2_related_work.tex
\section{Related Work}
\label{sec:RW}

\subsection{Machine Unlearning Methods}

Machine unlearning aims to remove targeted data or knowledge from a trained model without retraining it from scratch~\cite{ginart2019making,guo2020certified,bourtoule2021machine}. In large language models, factual knowledge is distributed across model parameters, making it challenging to erase targeted knowledge while preserving unrelated capabilities. Representative unlearning methods include GA~\cite{jang2023knowledge}, DPO~\cite{rafailov2023dpo}, NPO~\cite{zhang2024npo}, and RMU~\cite{li2024wmdp}. However, fine-tuning may alter how knowledge is retrieved rather than eliminate the underlying knowledge, allowing it to be recovered through paraphrased or adversarial queries~\cite{yao2024machine,barbulescu2024each,xu2025relearn,hong2024dissecting,patil2024sensitive,joshi2024robust}. This limitation is particularly pronounced in entity-level unlearning, where target entities are embedded in interconnected networks of factual relations~\cite{ma2025entity}.

\begin{figure*}[t]
    \centering
    \includegraphics[width=0.95\textwidth]{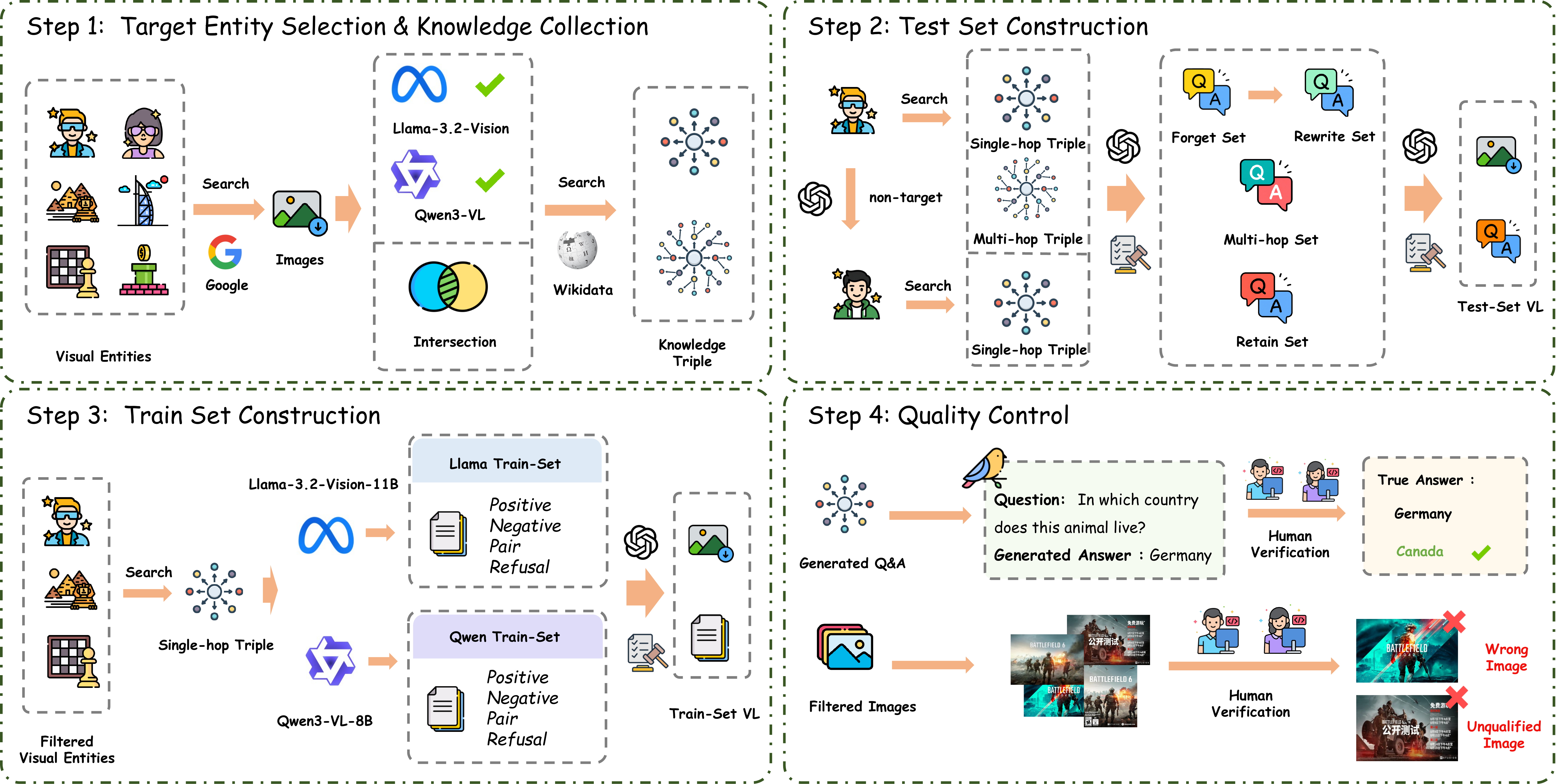}
    \caption{The four-stage construction pipeline of UNLINK-VL.}
    \label{fig:data_construction_pipeline}
    \vspace{-3mm}
\end{figure*}

\subsection{Machine Unlearning Benchmarks}

Existing machine-unlearning benchmarks evaluate synthetic profiles, hazardous knowledge, and the trade-off between forgetting efficacy and model utility~\cite{maini2024tofu,li2024wmdp,shi2025muse}. RWKU extends this line of research to a practical post-hoc setting, in which a model must forget previously acquired knowledge about public figures without access to the original forget corpora~\cite{jin2024rwku}. A key challenge in this setting is to distinguish genuine forgetting from weak prior knowledge, as a model that initially knows little about a target may appear to have been successfully unlearned~\cite{kwon2026before}.
Consequently, unlearning effects achieved in one modality may not transfer to the other modality~\cite{li2026mipeditor}. Existing benchmarks focus on different aspects of multimodal unlearning. MLLMU-Bench evaluates the forgetting of synthetic personal knowledge in text-only and text-image settings~\cite{liu2025mllmubench}. FIU-Bench studies identity-related knowledge associated with synthetic facial images and tests whether such knowledge can be recovered through privacy attacks~\cite{ma2025fiubench}. Other benchmarks extend multimodal unlearning to visual concepts, misinformation, and sensitive associations~\cite{li2024siu,zheng2025offside,selvassala2026salmubench}. Collectively, these studies show that unlearning effectiveness depends on what is forgotten, how it is represented, and how it is evaluated.
Among existing benchmarks, UMU-Bench and PPU-Bench are most closely related to our setting. UMU-Bench evaluates unimodal, multimodal, and mixed-modality unlearning using synthetic profiles, with a focus on modality alignment after unlearning~\cite{wang2025umubench}. By contrast, PPU-Bench focuses on knowledge that models already possess about public figures and evaluates complete, selective, and personalized unlearning through text-only and text-image QA~\cite{guang2026ppubench}. UNLINK-VL differs from both benchmarks by focusing on real-world knowledge and systematically examining whether unlearning effects induced in one modality transfer to another. Appendix D provides a detailed comparison with representative text-only and multimodal unlearning benchmarks.

%% file: Sec/3_datasets.tex
\section{UNLINK-VL Benchmark}
\label{sec:method}

\subsection{Task Definition}

\noindent\textbf{Knowledge unlearning.}
Suppose that an already trained model $M_{\mathrm{vanilla}}$ is asked to forget a set of target entities $E_{\mathrm{forget}}$. The goal is simple: after unlearning, the model should no longer answer questions about the target knowledge, but should still answer questions about unrelated knowledge.

Conceptually, the original training set $D$ can be divided into the data to be forgotten, $D_{\mathrm{forget}}$, and all remaining data, $D_{\mathrm{rest}}$:
\begin{equation}
\begin{gathered}
    D = D_{\mathrm{forget}}\cup D_{\mathrm{rest}},
    \qquad D_{\mathrm{forget}}\cap D_{\mathrm{rest}}=\varnothing,\\
    M_{\mathrm{unlearn}}
    = A(M_{\mathrm{vanilla}},D_{\mathrm{forget}}).
\end{gathered}
\label{eq:unlearning_definition}
\end{equation}
Here, $A$ is the unlearning algorithm. The ideal solution would remove $D_{\mathrm{forget}}$ and retrain from scratch on $D_{\mathrm{rest}}$; $A$ instead updates $M_{\mathrm{vanilla}}$ directly to obtain $M_{\mathrm{unlearn}}$~\cite{ginart2019making,guo2020certified,bourtoule2021machine}. In our practical post-hoc setting~\cite{jin2024rwku}, the original $D_{\mathrm{forget}}$ and $D_{\mathrm{rest}}$ are unavailable. We therefore construct a compact $D_{\mathrm{forget}}$ from public facts about $E_{\mathrm{forget}}$.

\noindent\textbf{Cross-modal knowledge unlearning.}
A fact is represented as a triple $k=(E,r,o)$, where $E$ is the
subject entity, $r$ is the relation, and $o$ is the correct answer.
For example, the fact
\begin{equation}
k =
(\mathrm{Taylor\ Swift},\ \mathrm{place\ of\ birth},\
 \mathrm{West\ Reading})
\end{equation}
states that Taylor Swift was born in West Reading. 

For each fact $(E,r,o)$, we construct two semantically equivalent
questions that differ only in how the subject entity is identified:
\begin{equation}
\begin{aligned}
Q_{k,T} &= G_T(E,r), \\
Q_{k,M} &= \bigl(I_E,G_M(E,r)\bigr), \\
A_{k,T} &= A_{k,M}=o.
\end{aligned}
\end{equation}
Here, $G_T$ generates a text-only question that explicitly names $E$,
whereas $G_M$ generates an image-grounded question that refers to $E$
through the image $I_E$. Both questions have the same answer $o$.

Further details on the paired query construction and evaluation settings are provided in Appendix B.

Our training and evaluation settings are as follows:
    \begin{equation}
    T\!\rightarrow\!T,\qquad
    T\!\rightarrow\!M,\qquad
    M\!\rightarrow\!T,\qquad
    M\!\rightarrow\!M.
    \end{equation}
    In $a\!\rightarrow\!b$, $a$ and $b$ denote the unlearning and
    evaluation modalities, respectively. The diagonal settings
    ($T\!\rightarrow\!T$ and $M\!\rightarrow\!M$) are intra-modal,
    whereas the off-diagonal settings ($T\!\rightarrow\!M$ and
    $M\!\rightarrow\!T$) evaluate cross-modal transfer.

\subsection{Dataset Construction}
\label{sec:data_construction}

To support reliable cross-modal evaluation while avoiding prompt and image leakage, we construct separate unlearning and evaluation branches grounded in the same underlying facts, as illustrated by the four-stage pipeline in Figure~\ref{fig:data_construction_pipeline}. Our pipeline draws on RWKU's focus on real-world target entities accessible to the evaluated models and their neighboring knowledge~\cite{jin2024rwku}, while following MMKU-Bench's staged organization of knowledge collection, cross-modal data construction, and quality control~\cite{fu2026mmku}. By keeping the target knowledge aligned across data splits, this design enables controlled evaluation of both intra-modal forgetting and cross-modal transfer.

\begin{figure}[!t]
    \centering
    \includegraphics[width=0.85\columnwidth]{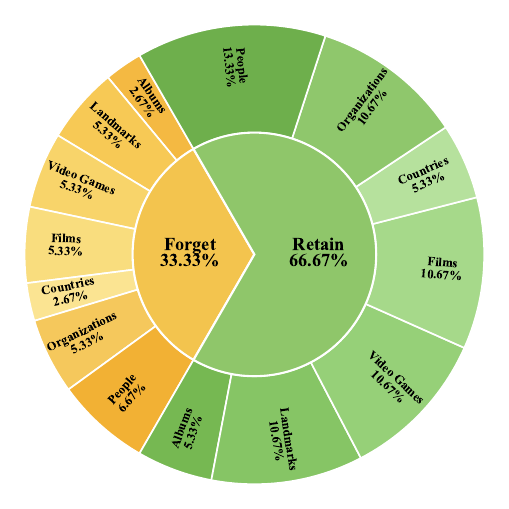}
    \caption{Entity-category distribution of the one-hop Forget and Retain
    evaluation sets.}
    \label{fig:forget_retain_categories}
    \vspace{-3mm}
\end{figure}

\noindent\textbf{Stage 1: Target entity selection and knowledge collection.}
To ensure that the unlearning targets are recognizable from both their names and images while avoiding image leakage between unlearning and evaluation, we begin with a broad pool of visually identifiable real-world entities spanning people, landmarks, organizations, countries, films, albums, games, and other categories. We retrieve candidate images through image search and conduct probing under both text-only and image-only settings on the Qwen3-VL and Llama-3.2-Vision model families to exclude entities unfamiliar to either family~\cite{kwon2026before}. The complete probing protocol and filtering criteria are provided in Appendix B. After filtering, 250 target entities remain. For each target, we collect two representative images, both of which independently satisfy the image-only probing criteria for both model families. To distinguish entity-level forgetting from image-specific effects, we use Image2 for multimodal unlearning and reserve Image1 for evaluation.

To assess not only forgetting but also its propagation through relational chains, we link each target entity to Wikidata~\cite{vrandecic2014wikidata} and retrieve both direct triples $(E,r,o)$ and valid two-hop paths $(E,r_1,E'),(E',r_2,o)$~\cite{zhong2023mquake}. We further identify semantically related non-target entities for retention evaluation~\cite{cohen2024ripple}. Each candidate entity and its associated image undergo the same text-only and image-only probing procedure, and only those recognized by both model families in both settings are retained. These entities constitute $E_{\text{retain}}$.

\noindent\textbf{Stage 2: Test-set construction.}
In order to separately evaluate forgetting, robustness to paraphrasing, recoverability through relational composition, and preservation of non-target knowledge, we organize the collected facts into four complementary evaluation subsets. The Forget Set consists of one-hop facts and measures whether the target knowledge is successfully forgotten. The Rewrite Set consists of GPT-4o-generated paraphrases of the Forget Set questions to evaluate robustness to alternative phrasings. The Multi-Hop Set is constructed from valid two-hop paths and assesses whether the target knowledge can still be recovered through relational composition. Finally, the Retain Set contains one-hop facts about non-target entities and evaluates whether such knowledge is preserved after unlearning.

GPT-4o converts the structured facts into clear, unambiguous questions with short answers grounded in the source triples. To guarantee that input modality acts as the sole variable in cross-modal comparisons, we instantiate each evaluation item as a fact-aligned modality pair. In the text-only version, the subject entity is named explicitly; in the text-image version, the question is paired with an image, and the entity name is replaced with an image-grounded referring expression, while the queried relation and answer remain unchanged. Thus, the two versions differ only in how the subject entity is accessed.

The complete prompts used for KG-to-QA generation, semantic rewriting, and cross-modal conversion are provided in Appendix C.

\noindent\textbf{Stage 3: Train-set construction.}
Since the pretrained datasets are unavailable, we construct model-specific synthetic training data from triples associated with the target entities retained after probing. Qwen3-VL-8B and Llama-3.2-Vision-11B each generate the forget corpus for their corresponding model family.

To enable a controlled comparison between text-only and multimodal unlearning, we construct fact-aligned text-image counterparts for all text-only unlearning instances in the same manner as the evaluation pairs. Dataset statistics are summarized in Table~\ref{tab:dataset_statistics} and Figure~\ref{fig:forget_retain_categories}. 

\noindent\textbf{Stage 4: Quality control.}
To prevent data-construction artifacts from being mistaken for unlearning effects and undermining the validity of the benchmark, we first apply automated screening to remove malformed, duplicate, ambiguous, or inconsistent samples, as well as unusable images. Human reviewers then correct unintended errors introduced during triple-to-QA conversion and verify the accuracy of image selection and image--entity alignment. Appendix E details the review unit, checklist, correction procedure, and final verification pass.

% The first table in the page layout is Table 1.
% \setcounter{table}{0}

\input{Tab/qa_statistics}
\input{Tab/qwen3vl8b_results}

% \FloatBarrier
\subsection{Unlearning Methods}
\label{sec:training_data}

We evaluate four representative methods. GA lowers the likelihood of the correct target response $y^{+}$~\cite{jang2023knowledge}; DPO ranks the counterfactual response $y^{-}$ above the faithful response $y^{+}$ relative to the original model~\cite{rafailov2023dpo}; NPO suppresses $y^{+}$ without requiring a preferred completion~\cite{zhang2024npo}; and RT applies supervised fine-tuning to $y^{\mathrm{ref}}$. All four methods are implemented with both full-parameter fine-tuning and LoRA.
For each target fact, the training data provide four forms of supervision: a faithful positive response $y^{+}$, a fluent counterfactual negative response $y^{-}$, a preference pair $(y^{-},y^{+})$, and a standardized refusal response $y^{\mathrm{ref}}$. Here, $y^{-}$ is the preferred completion and $y^{+}$ is the rejected completion in the preference pair. Positive and negative responses are generated and filtered separately for each target model, and their matched combination forms the paired supervision used by DPO. The same fact is instantiated across different modalities and supervision schemes, enabling controlled comparisons of GA, DPO, NPO, and RT while keeping the target knowledge fixed.

% \vspace{-6mm}
\subsection{Metrics}
\label{sec:metrics}

We evaluate knowledge unlearning along two standard axes: forgetting efficacy and retained utility~\cite{maini2024tofu,zhang2024npo}. We report answer accuracy on each subset. Lower accuracy on the Forget, Multi-Hop, and Rewrite Sets indicates stronger forgetting, whereas higher accuracy on the Retain Set indicates better preservation of neighboring knowledge. Besides, general capabilities are reported using the official metric of each external benchmark.

% \FloatBarrier

%% file: Tab/qa_statistics.tex
\begin{table}[t]
    \centering
    \small
    \setlength{\tabcolsep}{6pt}
    \renewcommand{\arraystretch}{1.08}
    \begin{tabular}{@{}lr@{}}
        \toprule
        \textbf{Statistics} & \textbf{Number} \\
        \midrule
        \multicolumn{2}{@{}l}{\textit{Training set (per model family)}} \\
        \quad Faithful samples      & 13,173 \\       
        \quad Counterfactual samples          & 13,173 \\
        \quad Text preference pairs           & 13,173 \\
        \quad Refusal samples       & 10,000 \\
        \midrule
        \multicolumn{2}{@{}l}{\textit{Test set}} \\
        \quad Forget Set   & 11906 \\
        \quad Multi-Hop Set & 3398  \\
        \quad Rewrite Set   & 11906 \\
        \quad Retain Set    & 20351 \\
        \midrule
        \multicolumn{2}{@{}l}{\textit{Entity}} \\
        \quad Forget & 250 \\
        \quad Retain & 500 \\
        \bottomrule
    \end{tabular}
    \caption{Key statistics of UNLINK-VL.}
    \label{tab:dataset_statistics}
    \vspace{-6mm}
\end{table}

%% file: Tab/qwen3vl8b_results.tex
\begin{table*}[t]
\centering
\scriptsize
\setlength{\tabcolsep}{2.6pt}
\renewcommand{\arraystretch}{1.08}
\resizebox{0.93\textwidth}{!}{%
\begin{tabular}{l*{16}{c}}
\toprule
& \multicolumn{4}{c}{$T\!\rightarrow\!M$} & \multicolumn{4}{c}{$M\!\rightarrow\!M$} & \multicolumn{4}{c}{$M\!\rightarrow\!T$} & \multicolumn{4}{c}{$T\!\rightarrow\!T$} \\
\cmidrule(lr){2-5}\cmidrule(lr){6-9}\cmidrule(lr){10-13}\cmidrule(lr){14-17}
\textbf{Method} & F$\downarrow$ & R$\uparrow$ & MH$\downarrow$ & RW$\downarrow$ & F$\downarrow$ & R$\uparrow$ & MH$\downarrow$ & RW$\downarrow$ & F$\downarrow$ & R$\uparrow$ & MH$\downarrow$ & RW$\downarrow$ & F$\downarrow$ & R$\uparrow$ & MH$\downarrow$ & RW$\downarrow$ \\
\midrule
Vanilla    & 82.6 & 80.8 & 66.9 & 79.4 & 82.6 & 80.8 & 66.9 & 79.4 & 85.1 & 82.3 & 69.7 & 82.0 & 85.1 & 82.3 & 69.7 & 82.0 \\
\midrule
GA (Full)  & 74.2 & 63.8 & 62.8 & 72.5 & 52.4 & 61.5 & 54.6 & 57.1 & 66.8 & 65.0 & 60.2 & 68.9 & 55.1 & 63.2 & 56.8 & 59.5 \\
GA (LoRA)  & 79.0 & 72.8 & 65.2 & 76.5 & 65.3 & 70.6 & 60.7 & 67.2 & 73.5 & 73.7 & 64.3 & 74.9 & 67.2 & 71.4 & 62.8 & 68.8 \\
DPO (Full) & 75.8 & 72.9 & 63.0 & 73.6 & 56.0 & 70.8 & 55.8 & 59.8 & 68.4 & 73.8 & 60.9 & 70.4 & 58.9 & 71.6 & 57.4 & 61.6 \\
DPO (LoRA) & 79.6 & 78.7 & 65.0 & 77.0 & 67.1 & 77.9 & 61.2 & 68.6 & 74.6 & 79.2 & 64.1 & 75.4 & 68.5 & 77.1 & 62.0 & 69.2 \\
NPO (Full) & 71.0 & 68.9 & 61.8 & 69.4 & 45.2 & 66.1 & 50.7 & 53.8 & 62.9 & 69.8 & 57.8 & 66.3 & 49.8 & 67.4 & 53.1 & 55.6 \\
NPO (LoRA) & 75.4 & 75.5 & 63.6 & 73.4 & 58.4 & 73.7 & 56.9 & 61.2 & 69.2 & 76.6 & 61.2 & 71.0 & 62.0 & 73.8 & 58.5 & 63.5 \\
RT (Full)  & 72.6 & 71.7 & 62.1 & 70.5 & 48.8 & 69.9 & 52.3 & 55.4 & 64.3 & 72.4 & 58.5 & 67.4 & 52.9 & 70.1 & 54.2 & 57.2 \\
RT (LoRA)  & 76.8 & 77.6 & 64.1 & 74.5 & 60.5 & 76.2 & 58.1 & 63.1 & 70.8 & 78.3 & 61.8 & 72.6 & 63.8 & 75.4 & 59.4 & 65.2 \\
\bottomrule
\end{tabular}}
\caption{The results of Qwen3-VL-8B on  UNLINK-VL. $T$ and $M$ denote text-only and multimodal inputs, respectively.
% and $X\!\rightarrow\!Y$ indicates that the model is unlearned using modality $X$ and evaluated using modality $Y$.
F, R, MH, and RW denote the Forget, Retain, Multi-Hop, and Rewrite scores,
respectively. 
% $\downarrow$ and $\uparrow$ indicate that lower and higher
% values are better, respectively.
}
\label{tab:qwen3vl8b_results}
\vspace{-3mm}
\end{table*}

%% file: Sec/4_exp.tex
\section{Experiments}
\label{sec:exp}

\setcounter{table}{2}
\input{Tab/general_capability}      % Table 3
\input{Tab/qwen3vl32b_results}      % Table 4

\subsection{Experimental Setup}

\noindent\textbf{Models.}
We evaluate UNLINK-VL on Qwen3-VL-8B, Qwen3-VL-32B, and Llama-3.2-Vision-11B. The two model families use different vision--language integration mechanisms, making them suitable for testing whether cross-modal unlearning patterns generalize across architectures. The main paper reports representative Qwen3-VL-8B results and a LoRA-based 8B--32B scale comparison, while the consolidated Llama-3.2-Vision-11B results are deferred to Appendix A.

\noindent\textbf{Implementation.}
All models are trained in \texttt{bf16} for three epochs with a maximum sequence length of 1,024, a per-device batch size of 1, and four gradient-accumulation steps. We use a cosine learning-rate schedule with 20 warm-up steps; the learning rate is selected for each objective and backbone. LoRA uses rank $r=32$, scaling factor $\alpha=64$, and targets all supported linear modules. Full-parameter training optionally uses DeepSpeed ZeRO-2~\cite{rajbhandari2020zero}. The 8B and 11B LoRA runs use two NVIDIA A100-80GB GPUs, and their full-parameter runs use four; the 32B LoRA runs use four A100-80GB GPUs. We train separate text-only and multimodal unlearning variants for each method and evaluate them with the same standalone pipeline over the Forget, Multi-Hop, Rewrite, and Retain Sets in both modalities.

\begin{figure*}[t]
    \centering
    \includegraphics[width=0.95 \textwidth]{3.pdf}
    \caption{Case Study}
    \label{fig:case_study}
    \vspace{-3mm}
\end{figure*}

\subsection{Main Results}

For compact presentation, Table~\ref{tab:qwen3vl8b_results} merges the four unlearning--evaluation configurations for Qwen3-VL-8B. Each configuration contains Forget (F), Retain (R), Multi-Hop (MH), and Rewrite (RW) accuracy.

\textbf{(1) Intra-modal unlearning is generally more effective with multimodal inputs.}
Since text-only and text-image evaluations have different Vanilla baselines, we compare baseline-normalized Forget reductions. In six of eight method--update pairs, $M\rightarrow M$ yields a larger relative reduction than $T\rightarrow T$, by 1.3--3.8 percentage points. The two LoRA exceptions are GA, which is nearly tied (20.9 \% versus 21.0\%), and DPO, where $T\rightarrow T$ is slightly stronger (19.5\% versus 18.8\%). Thus, multimodal inputs generally strengthen intra-modal suppression, though not uniformly under LoRA.

\textbf{(2) Cross-modal forgetting remains incomplete and asymmetric.}
For every method and update regime, cross-modal evaluation yields a smaller baseline-normalized Forget reduction than the corresponding intra-modal evaluation. The relative $T\rightarrow M$ penalty ranges from 15.9 to 27.4 percentage points, whereas the $M\rightarrow T$ penalty ranges from 6.4 to 19.2 points. Moreover, the latter is smaller for all eight method--update pairs, averaging 12.2 points compared with 21.2 for $T\rightarrow M$. These normalized comparisons remove modality-specific Vanilla differences and confirm that multimodal unlearning transfers more effectively to text-only queries than text-only unlearning transfers to text-image queries, so same-modality evaluation can overestimate forgetting.

\textbf{(3) NPO forgets most, DPO preserves most, and RT provides the best balance.}
Under the same update and unlearning--evaluation configuration, NPO always achieves the lowest Forget accuracy and therefore removes the most target knowledge. DPO always achieves the highest Retain accuracy and causes the least damage to neighboring knowledge. RT forgets more target knowledge than DPO while preserving more neighboring knowledge than NPO, giving the best overall balance. GA performs less favorably: it forgets less than NPO and RT while usually obtaining the lowest Retain accuracy.

\textbf{(4) Neighboring knowledge is largely preserved, but forgotten knowledge remains recoverable.}
Across all trained variants and modality configurations, Retain accuracy remains between 61.5 and 79.2, showing that much neighboring knowledge survives, although some collateral damage remains. However, accuracy on the Multi-Hop and Rewrite Sets remains substantial, ranging from 50.7 to 65.2 and from 53.8 to 77.0, respectively. Thus, even when a model fails to answer a direct Forget Set question, it may recover the same target knowledge through relational reasoning or a semantically equivalent question. Effective unlearning needs to preserve neighboring knowledge while preventing such recovery.

\subsection{Analysis Experiments}

\textbf{General-Capability Evaluation.}
We evaluate unrelated utility on the ten benchmarks listed in
Table~\ref{tab:general_capability} using their official
protocols~\cite{fu2023mme,liu2024mmbench,li2024seedbench,
liu2024ocrbench,lu2022scienceqa,qian2025miabench,
lu2024mathvista,wang2024mathvision,li2023pope,
guan2024hallusionbench}.

Table~\ref{tab:general_capability} shows that LoRA generally preserves broad
utility, although losses are task and objective dependent. MME stays within
$-0.57\%$ to $+0.58\%$ of Vanilla, while OCRBench and POPE change by at most
0.7 and 0.51 points. The largest drops occur in scientific and mathematical
reasoning: GA reduces ScienceQA from 74.27 to 68.52 (text) and 68.47
(multimodal), and its text variant lowers MathVista from 78.9 to 76.2. DPO
with multimodal unlearning remains near or above Vanilla on several broad and
instruction-following benchmarks. Neither unlearning modality is uniformly
better; small gains are treated as retained capability or evaluation
variation. Thus, general utility is largely preserved, but the few pronounced
losses must be considered alongside forgetting and locality.

\textbf{Full-Parameter Fine-Tuning versus LoRA.} The matched runs reveal a clear forgetting-locality trade-off. Relative to LoRA, full-parameter tuning lowers Forget accuracy by 3.8--13.2 points, but also lowers
Retain accuracy by 5.3--9.1 points; it similarly strengthens Multi-Hop and
Rewrite suppression in every setting. NPO (Full) suppresses most strongly,
DPO (LoRA) preserves the most neighboring knowledge, and RT (LoRA) provides
the best balance between them. LoRA also better maintains the broad utility
reported in Table~\ref{tab:general_capability}. Full tuning is therefore
preferable when suppression is paramount, whereas LoRA is the more
conservative choice when locality and general utility matter.

\textbf{Model-Size Effects.} To account for different Vanilla baselines, we compare baseline-normalized changes. Across matched LoRA runs, the 32B Forget reductions are 0.02--2.76 percentage points smaller, indicating greater resistance to fixed-rank LoRA. Although absolute Retain accuracy is 2.5--2.8 points higher, proportional Retain loss is 0.05--0.85 points larger, showing that the absolute advantage mainly reflects the stronger baseline. Method ordering remains stable: NPO suppresses most, DPO preserves most, and RT offers a favorable balance. Cross-modal asymmetry persists: the normalized $T\rightarrow M$ penalty is 14.0--16.2 points, versus 6.3--10.3 for $M\rightarrow T$.

\subsection{Case Study}
\label{sec:qualitative_case_study}

Figure~\ref{fig:case_study} presents a representative example of the paired text-only and text-image instances used for unlearning and evaluation, all constructed from the same underlying fact. It further illustrates the four unlearning--evaluation settings and the corresponding model outputs under different unlearning methods and tuning strategies, enabling a direct comparison of intra-modal and cross-modal transfer as well as full-parameter and LoRA-based adaptation. 

%% file: Tab/general_capability.tex
\begin{table*}[t]
\newcommand{\gcdown}[2]{\begingroup\setlength{\fboxsep}{1.2pt}\colorbox{red!#1}{\strut #2}\endgroup}
\centering
\scriptsize
\setlength{\tabcolsep}{3.5pt}
\resizebox{0.95\textwidth}{!}{%
\begin{tabular}{llcccccccccc}
\toprule
& \textbf{Method} & \multicolumn{3}{c}{\textbf{Comprehensive}} & \textbf{OCR} & \textbf{Multidisciplinary} & \textbf{Instruction} & \multicolumn{2}{c}{\textbf{Mathematical}} & \multicolumn{2}{c}{\textbf{Hallucination}} \\
\cmidrule(lr){3-5}\cmidrule(lr){6-6}\cmidrule(lr){7-7}\cmidrule(lr){8-8}\cmidrule(lr){9-10}\cmidrule(lr){11-12}
& & MME$\uparrow$ & MMBench$\uparrow$ & SEED$\uparrow$ & OCRBench$\uparrow$ & ScienceQA$\uparrow$ & MIA-Bench$\uparrow$ & MathVista$\uparrow$ & MathVision$\uparrow$ & POPE$\uparrow$ & HallusionBench$\uparrow$ \\
\midrule
& Vanilla & 1734.77 & 75.43 & 67.24 & 90.9 & 74.27 & 93.83 & 78.9 & 49.01 & 88.44 & 73.50 \\
\midrule
\multirow{4}{*}{$T$} & GA (LoRA)
& \gcdown{11}{1732.25}
& \gcdown{21}{73.80}
& \gcdown{28}{64.78}
& 91.2
& \gcdown{49}{68.52}
& \gcdown{12}{93.45}
& \gcdown{27}{76.2}
& 53.29
& \gcdown{12}{88.05}
& \gcdown{14}{72.87} \\
& DPO (LoRA)
& 1736.63
& \gcdown{11}{75.34}
& \gcdown{20}{65.88}
& \gcdown{12}{90.5}
& \gcdown{23}{72.38}
& 94.04
& \gcdown{14}{78.3}
& 50.00
& \gcdown{12}{88.13}
& 74.13 \\
& NPO (LoRA)
& 1737.43
& \gcdown{12}{75.17}
& \gcdown{26}{65.04}
& \gcdown{12}{90.5}
& \gcdown{38}{70.05}
& 94.21
& \gcdown{19}{77.5}
& 51.97
& \gcdown{12}{88.03}
& 73.92 \\
& RT (LoRA)
& \gcdown{12}{1729.35}
& 76.03
& \gcdown{10}{67.19}
& 91.0
& \gcdown{14}{73.62}
& \gcdown{10}{93.76}
& \gcdown{19}{77.5}
& 52.63
& \gcdown{13}{87.95}
& \gcdown{15}{72.77} \\
\midrule
\multirow{4}{*}{$M$} & GA (LoRA)
& \gcdown{13}{1724.85}
& \gcdown{12}{75.09}
& \gcdown{19}{66.01}
& 91.1
& \gcdown{49}{68.47}
& 94.23
& \gcdown{14}{78.2}
& \gcdown{13}{48.68}
& \gcdown{12}{88.04}
& \gcdown{12}{73.19} \\
& DPO (LoRA)
& 1744.75
& 76.46
& 68.07
& \gcdown{12}{90.5}
& 75.76
& 94.52
& \gcdown{14}{78.3}
& 52.30
& \gcdown{12}{88.16}
& \gcdown{14}{72.98} \\
& NPO (LoRA)
& 1737.52
& \gcdown{13}{75.00}
& \gcdown{19}{66.10}
& 91.6
& \gcdown{29}{71.39}
& 93.97
& \gcdown{15}{78.1}
& 51.32
& \gcdown{13}{87.97}
& 74.34 \\
& RT (LoRA)
& 1742.37
& 75.60
& \gcdown{12}{66.93}
& 90.9
& \gcdown{18}{73.03}
& 94.30
& \gcdown{25}{76.5}
& 51.32
& \gcdown{13}{87.93}
& 73.71 \\
\bottomrule
\end{tabular}}
\caption{General-capability results for Qwen3-VL-8B after unlearning.}
\label{tab:general_capability}
\vspace{-6mm}
\end{table*}

%% file: Tab/qwen3vl32b_results.tex
\begin{table*}[t]
\centering
\scriptsize
\setlength{\tabcolsep}{2.6pt}
\renewcommand{\arraystretch}{1.08}
\resizebox{0.9\textwidth}{!}{%
\begin{tabular}{l*{16}{c}}
\toprule
& \multicolumn{4}{c}{$T\!\rightarrow\!M$} & \multicolumn{4}{c}{$M\!\rightarrow\!M$} & \multicolumn{4}{c}{$M\!\rightarrow\!T$} & \multicolumn{4}{c}{$T\!\rightarrow\!T$} \\
\cmidrule(lr){2-5}\cmidrule(lr){6-9}\cmidrule(lr){10-13}\cmidrule(lr){14-17}
\textbf{Method} & F$\downarrow$ & R$\uparrow$ & MH$\downarrow$ & RW$\downarrow$ & F$\downarrow$ & R$\uparrow$ & MH$\downarrow$ & RW$\downarrow$ & F$\downarrow$ & R$\uparrow$ & MH$\downarrow$ & RW$\downarrow$ & F$\downarrow$ & R$\uparrow$ & MH$\downarrow$ & RW$\downarrow$ \\
\midrule
Vanilla    & 85.9 & 84.1 & 71.8 & 83.2 & 85.9 & 84.1 & 71.8 & 83.2 & 88.0 & 85.4 & 74.5 & 85.6 & 88.0 & 85.4 & 74.5 & 85.6 \\
\midrule
GA   & 82.3 & 75.3 & 67.2 & 79.2 & 69.3 & 73.1 & 63.1 & 70.1 & 77.2 & 76.2 & 66.7 & 77.8 & 71.2 & 73.9 & 65.2 & 71.7 \\
DPO  & 82.8 & 81.2 & 67.3 & 79.8 & 71.2 & 80.5 & 63.7 & 71.5 & 78.5 & 82.0 & 66.6 & 78.5 & 72.5 & 79.9 & 64.6 & 72.3 \\
NPO  & 78.8 & 78.1 & 66.0 & 76.5 & 62.8 & 76.4 & 59.4 & 64.5 & 73.4 & 79.3 & 63.8 & 74.2 & 66.5 & 76.5 & 61.2 & 66.7 \\
RT   & 80.2 & 80.2 & 66.6 & 77.5 & 64.9 & 78.9 & 60.6 & 66.4 & 75.0 & 81.0 & 64.4 & 75.5 & 68.4 & 78.2 & 62.2 & 68.5 \\
\bottomrule
\end{tabular}}
\caption{Results on UNLINK-VL using Qwen3-VL-32B with LoRA.}
\label{tab:qwen3vl32b_results}
% \vspace{-3mm}
\end{table*}

%% file: Sec/5_conclusion.tex
\section{Conclusion}
\label{sec:con}
We introduced UNLINK-VL, a real-world benchmark for cross-modal knowledge unlearning in Vision-Language Models. UNLINK-VL independently controls the modalities used for unlearning and evaluation, separating intra-modal forgetting from cross-modal transfer. It grounds model-known, visually identifiable entities in one-hop facts and two-hop Wikidata paths and organizes paired text-only and text-image probes into the Forget, Multi-Hop, Rewrite, and Retain Sets. Together, these components evaluate forgetting, relational propagation, robustness, neighboring-knowledge preservation, and general utility. The experiments reveal a pronounced transfer asymmetry: multimodal unlearning transfers more effectively to text-only evaluation than text-only unlearning transfers to text-image evaluation. At the same time, the evaluated methods largely preserve general capabilities. These findings show that text-only evaluation can substantially overestimate knowledge unlearning in VLMs and underscore the need for cross-modal unlearning and evaluation.